\documentclass[runningheads]{llncs}
\usepackage[T1]{fontenc}
\usepackage{graphicx}
\usepackage{booktabs}
\usepackage[misc]{ifsym}
\newcommand{\corr}{(\Letter)}
\usepackage{xcolor} % Needed for defining and using colors
\definecolor{paleorange}{RGB}{255, 229, 204} % A soft, light orange
\definecolor{palegrey}{RGB}{230, 230, 230}   % A light grey
\definecolor{paleblue}{RGB}{204, 229, 255}   % A soft, light blue
\definecolor{palegreen}{RGB}{204, 255, 204}  % A light, minty green
\definecolor{palered}{RGB}{255, 204, 204}    % A soft, light red

\usepackage{multirow}
\usepackage{silence}
\usepackage{mathtools}
\usepackage{booktabs} 
\usepackage{tikz}
\usetikzlibrary{fit}
\usepackage{amssymb}
\usepackage[framemethod=TikZ]{mdframed}
\usepackage{stmaryrd}
\usepackage{enumerate} 
\usepackage{colortbl}
\usepackage{bbm}
\usepackage{hyperref}
\usetikzlibrary{calc, positioning, arrows.meta, shapes.geometric, backgrounds}

\begin{document}
%\title{Fine-Tuning is All You Need: Compact Models Can Outperform GPT's Classification Abilities}
\title{Optimizing Performance: How Compact Models Match or Exceed GPT's Classification Capabilities through Fine-Tuning}

\titlerunning{Compact Models Can Outperform GPT's Classification Abilities}

%N.B.: Author information (both in the \author{} and \authorrunning{} command) should only be present in the Camera-Ready Version of your paper. The version that you initially submit for review, ought to be double-blind. So, when initially submitting your paper, use:

\author{Baptiste Lefort\inst{1,2} \corr \and
Eric Benhamou\inst{1,3} \and
Jean-Jacques Ohana\inst{1} \and David Saltiel \inst{1} \and Beatrice Guez\inst{1}}

\institute{Ai For Alpha, France, \email{baptiste.lefort@aiforalpha.com} \and CentraleSupelec, France \and Paris Dauphine - PSL, France}

\authorrunning{Lefort et al.}
\titlerunning{Compact Models Can Match or Exceed GPT's Sentiment Clasification}

\maketitle              % typeset the header of the contribution

\begin{abstract}
In this paper, we demonstrate that non-generative, small-sized models such as FinBERT and FinDRoBERTa, when fine-tuned, can outperform GPT-3.5 and GPT-4 models in zero-shot learning settings in sentiment analysis for financial news. These fine-tuned models show comparable results to GPT-3.5 when it is fine-tuned on the task of determining market sentiment from daily financial news summaries sourced from Bloomberg. To fine-tune and compare these models, we created a novel database, which  assigns a market score to each piece of news without human interpretation bias, systematically identifying the mentioned companies and analyzing whether their stocks have gone up, down, or remained neutral. Furthermore, the paper shows that the assumptions of Condorcet's Jury Theorem do not hold suggesting that fine-tuned small models are not independent of the fine-tuned GPT models, indicating behavioral similarities. Lastly, the resulted fine-tuned models are made publicly available on HuggingFace, providing a resource for further research in financial sentiment analysis and text classification.

\end{abstract}

\section{Introduction}\label{sec:intro}
Natural Language Processing (NLP) has shown a significant improvement in the task of sentiment analysis \cite{araci2019finbert,yang2020finbert,sohangir2018big,day2016deep,wu2023bloomberggpt,yang2023fingpt}. The Large Language Models are the most proficient models at the time. Their efficiency in interpreting complex patterns and processing an extensive context \cite{chen2023extending} is particularly useful in the financial sector. In the past, NLP models faced learning problems because they could only examine small amounts of data at a time, and there was little training data. Also, they could not process advanced concepts because they did not have enough parameters \cite{day2016deep}. Prompt engineering in large language models (LLMs) is linked to large context windows. These models react strongly to input prompts. Various methods, like the Chain-of-Thought, have shown to greatly improve these models \cite{wei2023chainofthought}. Using strategies similar to human problem-solving can boost the model's effectiveness. The technique of few-shot prompting is also beneficial in enhancing performance on diverse tasks in LLMs, as indicated in \cite{zhang-etal-2022-prompt-based}. The large quantity of parameters in the model facilitates its adjustment to excel in particular tasks. \\

New developments in NLP have made it better at understanding and categorizing text sentiment as "positive", "negative", or "indecisive". This is often used in finance, where it helps make sense of complex textual and numerical data \cite{9142175}. This field has unique challenges \cite{dumiter2023impact,briere2023stock}. For instance, financial news is often brief, lacks context, and can be difficult to interpret. Additionally, it combines text and numbers, which provides a complete picture of financial events but can add to its complexity. The news quickly loses relevance as it often discusses past events, and the exact impact is not always immediately clear. Financial data appears as short texts like headlines or tweets, making it hard for humans to interpret \cite{malo2013good}. One of the first challenge seems to build a reliable dataset. The financial models are currently trained with human-based annotated dataset \cite{Malo2014GoodDO}. The fine-tuned LLMs are evaluated based on this data. Even if they reach significant performances, there are no guarantees that the classification is correct from a financial point of view. This complex data provides an excellent opportunity to improve sentiment analysis understanding. Accordingly, we have developed a method to enhance the understanding of sentiment classification by LLMs when dealing with these short, complex texts.

\paragraph{Contributions:}
The contribution of this paper is manifold:
\begin{enumerate}
    \item \textbf{A market-based dataset.} We propose a market-based dataset built on real, objective events. The process we use has enabled us to annotate a large amount of data, from which we have been able to accurately refine and evaluate state-of-the-art LLMs for this 3-class classification task. 
    \item \textbf{Accurate fine-tuning and evaluation of the LLMs.} We fine-tune FinBERT, FinDRoBERTa and GPT-3.5 on the market-based dataset. We obtain an accurately trained model and evaluate them precisely on financial classification capability. 
    \item \textbf{LLMs performances reconsideration. } We observed comparable results, which led us to reconsider the size advantage of the GPT-3.5/4 model over more compact models. We propose a mathematical approach to detail our results.
    \item \textbf{Contributing to future research. } We are making our dataset \footnote{\url{https://huggingface.co/datasets/baptle/financial_headlines_market_based}} and fine-tuned models\footnote{\url{https://huggingface.co/baptle/FinBERT_market_based}, \url{https://huggingface.co/baptle/FinDROBERT_market_based}} open-source on the HuggingFace platform for reproducibility. We aim to provide this content to enable further work and improvements.
\end{enumerate}

\section{Related Works}\label{sec:related_works}
In the field of sentiment classification for financial texts, LLMs have been widely adopted, demonstrating high accuracy in real applications \cite{Hansen2023CanFedspeak,Cowen2023HowGPT,Korinek2023LanguageResearch,Lopez-Lira2023CanModels,Noy2023ExperimentalIntelligence,lefort2024chatgpt,zhao2024revolutionizing}. These studies collectively affirm the efficacy of LLMs in discerning sentiments in financial texts, marking a significant step forward in the application of artificial intelligence in finance.

The advent of FinBERT, a model specifically designed for financial text analysis through the process of fine-tuning LLMs on financial datasets, has led to significant progress in performance \cite{araci2019finbert}. This advancement underscores the potential of targeted model optimization to enhance accuracy and relevance in sector-specific applications.

Moreover, the introduction of niche datasets such as FinQA or financial-phrasebank has played a major role in refining the proficiency of LLMs in processing financial texts interspersed with numerical information \cite{chen2022finqa,Malo2014GoodDO}. These datasets train models to adeptly navigate the dual landscape of textual and numerical data, an essential capability for analyzing financial news that often melds narrative with figures.

The research into a more compact version of the BERT model, known as DistilBERT, has yielded optimistic results by demonstrating that reducing the number of parameters does not significantly compromise the model's ability to classify sentiments \cite{sanh2020distilbert}. This finding is crucial as it validates the efficiency of streamlined models in conducting sentiment analysis, particularly in the context of financial documents that typically present a complex blend of text and data.

The process of fine-tuning models has been notably successful in the analysis of complex financial documents, which frequently combine various types of inputs \cite{li2023multimodal}. This technique enhances the model's understanding and interpretation of the multifaceted information contained within these documents.

Furthermore, the integration of Retrieval Augmented Large Language Models has elevated the capacity of LLMs to conduct sentiment analysis by weaving in a broader contextual backdrop \cite{zhang2023enhancing}. This approach enriches the analysis by drawing upon external knowledge, which is particularly beneficial in the financial news sector where understanding the deeper, often unspoken implications of news is crucial.

The acquisition of extensive information, pertinent for grasping the sentiment in financial news, emerges as a significant hurdle, necessitating deep, contextual knowledge beyond the immediate content of the news articles \cite{kim2023context}. The challenge is compounded by the sheer volume of information that needs to be sifted through, as highlighted by \cite{Loukas_2023}.

The task of classifying news headlines for sentiment analysis is complicated by the inherent brevity of headlines, which often lack sufficient context. This scarcity of context poses a challenge for LLMs, hindering their ability to accurately infer the underlying sentiments \cite{zhang2023sentiment}. Despite these obstacles, some LLMs have shown a remarkable ability to interpret headlines effectively. \cite{lefort2024chatgpt} introduced an indicator that correlates the sentiments extracted from headlines with movements in equity markets, thereby attesting to GPT's efficiency in sentiment analysis even in limited contexts. This also validated the interest of leveraging textual data for financial applications.

In assessing the performance of LLMs in sentiment classification, comparisons are often drawn with annotations made by human experts \cite{briere2023stock,araci2019finbert}. However, this comparison is inherently bias, as it may not accurately reflect the real-time dynamics of the market. \cite{lewis2021retrievalaugmented}, have shown that LLMs exhibit a lower tendency to generate unfounded content ("hallucinate") when their analyses are grounded in factual data, emphasizing the importance of data veracity in enhancing the reliability of sentiment analysis outcomes.

\section{Data and Methodology}\label{sec:data_method}
\subsection{Data Collection}\label{subsec:data_collection}
We collected data from Bloomberg Market Wraps spanning 2010 to 2024. These news, created by financial experts, provide a condensed overview of daily financial events and are widely used in the industry. They offer valuable information for analysis as they cover all significant news. The extra benefit of using Bloomberg Market Wraps is to get information about all markets and regions. We want this because we aim to get the most general news and not just focus on a specific area or market. They are distributed through various channels like the Bloomberg network and web channels like Yahoo finance and Investing. We have collected over 3,700 comprehensive, high-quality daily news articles, which in turn summarize between 500 and 1,000 individual news items, giving a total number of scanned news items in excess of 2 million. However, instead of using individual news, we concentrate on Bloomberg daily market wraps to benefit from human expertise to identify key events and themes impacting financial markets, hence reducing noise in our news dataset. In addition, we used a two-step approach to eliminate noise and structure the data.

\subsection{Headline Generation}\label{subsec:headline_generation}
After collecting the daily news, we extracted headlines highlighting the day's most important information, enabling us to summarize the information effectively. This approach helped us concentrate on the vital aspects by filtering out minor details and condensing the news into brief sentences. We also get headlines that don't contradict each other. Additionally, it allowed us to gather more information by eliminating unnecessary noise and isolating the key facts in the headlines. The resulting headlines are both informative and useful for making investment decisions. Below is the prompt used to generate these headlines with GPT-4 model.

\vspace{1cm}

\begin{mdframed}[linecolor=black!20, innerleftmargin=10pt, innerrightmargin=10pt, innertopmargin=10pt, 
innerbottommargin=10pt, backgroundcolor=black!3, roundcorner=3pt]
\textit{You are provided with a financial text, and your task is to extract a list of headlines from it. Each headline must be informative and provide relevant insights for a financial market analyst. Ensure that each headline contains a single piece of information. List these headlines in the specified format, with each headline separated by a line break and without additional commentary. Format your list as follows:}
\begin{enumerate}
    \item \textit{Headline for Theme 1}
    \item \textit{Headline for Theme 2}
    \item \textit{...}
\end{enumerate}
\end{mdframed}

This two-step approach is valuable for improving model's classification performances, as it delivers meaningful and noise-free information \cite{lefort2024chatgpt}. At this point, we obtain a dataset of unlabeled short texts.

\section{A Market-Based Dataset For Evaluation}\label{sec:baseline_pres}
In this section, we propose a novel dataset for fine-tuning and evaluating the financial LLMs on the sentiment classification task. We introduce the need of a human-bias free dataset and detail the process followed for building it.

\subsection{The Need of a Market-Based Dataset}\label{sec:needofmarketbaseddataset}
To the best of our knowledge, sentiment classification models associated with financial news are trained and evaluated on datasets annotated by humans \cite{malo2013good}. These headlines are evaluated by a group of financial experts (or master's degree students) who assign  a label for each text. For example, the FinBERT model is trained on a dataset where "Three of the annotators were researchers and the remaining 13 annotators were master’s students" \cite{araci2019finbert}. However, human assessment of a market reaction can vary considerably. It depends on the natural abilities and experience of each individual. For some, an event may be perceived as positive, for others as negative. It's also possible for all the annotators to get it wrong and mislabel a financial headline. This disrupts model fine-tuning and evaluation. For this reason, we developed an human-bias free data set. The label must reflect the market's true behavior, not just the human annotator's assessment.

\subsection{Ticker Identification}
Firstly, we use GPT-4 to identify and assign a list of tickers associated with each headline. The model has demonstrated efficiency in accurately determining the involved entities in a text \cite{wang2023gptner}, \cite{wu2023bloomberggpt}. By doing so, we identify the markets which are the most likely to react to the news. Indeed, a headline may concern specific markets and is unlikely to provoke a reaction on all the markets at once. This step ensures that the associated sentiment reflects the most precisely the true market conditions. In this setup, it's necessary to note that the headline is considered to have an equal impact on all the related tickers. This does not affect the labelling of the data since we want to capture the macro effect of the headline. If the majority of the associated markets have the same common sentiment, then the macro sentiment will be correctly evaluated. \cite{lefort2024chatgpt} detailed in their work that it exists a macro impact of the news on the markets. It remains that even if it seems accurate after double-checking by financial experts, we are aware that the hypothesis whereby headlines are entirely responsible for market movements cannot be completely dismissed. Table \ref{tab:ticker_identification} provides a subset of our broader fifty thousands headlines, illustrating specific data entries and their associated financial tickers. The table aligns dates with corresponding news headlines and the related financial market tickers, providing a snapshot from January 4, 2010, to January 31, 2024. Entries include significant financial events, such as shifts in the dollar value, stock market highs, and oil price changes, etc.

\begin{table}[!htbp]
    \centering
    \caption{Sample of our large Dataset and corresponding list of Ticker(s)}
    \label{tab:ticker_identification}
    \resizebox{\textwidth}{!}{
    \begin{tabular}{llc}
        \toprule
        \bfseries Date & \bfseries Headline & \bfseries Tickers \\
        \midrule
        2010-01-04 & Dollar Slumps Amid Worldwide Manufacturing Improvement & [UUP, ACWI] \\
        \ldots & \ldots & \ldots \\
        2022-08-17 & S\&P 500 Rises 1\% to All-Time High, Treasuries Lose Gains & [GSPC, TNX] \\
        \ldots & \ldots & \ldots \\
        2024-01-31 &  West Texas Intermediate Crude Falls 1.3\% to \$76.83 a Barrel & [CL=F]\\
        \bottomrule
    \end{tabular}
    }
\end{table}

Table \ref{tab:ticker_distribution} presents the distribution of key tickers throughout the entire dataset. Notably, the predominant tickers belong to major equity markets and more specifically to the US market, showing its predominance in the equity financial world and its substantial media coverage and influence. 
\begin{table}[!htbp]
    \centering
    \caption{Distribution of the tickers over the dataset by region and major tickers. Note that only the most represented tickers are in the table and the percentage does not sum at 100\%.}
    \label{tab:ticker_distribution}
    \resizebox{\textwidth}{!}{
    \begin{tabular}{lllr}
\toprule
\textbf{Region} & \textbf{Ticker} & \textbf{Financial Name} & \textbf{Percentage (\%)} \\ \hline
\multirow{4}{*}{US} & SPY & S\&P 500 ETF Trust & 12.73 \\ \cline{2-4} 
 & GSPC & S\&P 500 Index & 8.92 \\ \cline{2-4} 
 & TNX & 10-Year Treasury Note Yield & 4.26 \\ \cline{2-4} 
 & UUP & Invesco DB US Dollar Index Bullish Fund & 0.97 \\ \hline

\multirow{4}{*}{Asia} & MCHI & iShares MSCI China ETF & 2.54 \\ \cline{2-4} 
 & N225 & Nikkei 225 Index & 1.19 \\ \cline{2-4} 
 & EWJ & iShares MSCI Japan ETF & 1.32 \\ \cline{2-4} 
 & EWA & iShares MSCI Australia ETF & 0.68 \\ \hline

\multirow{2}{*}{Europe} & EWG & iShares MSCI Germany ETF & 1.19 \\ \cline{2-4} 
 & VGK & Vanguard FTSE Europe ETF & 0.73 \\ \hline

 \multirow{1}{*}{Emerging Countries} & EEM & iShares MSCI Emerging Markets ETF & 2.07 \\ \hline

\multirow{3}{*}{Global} & ACWI & iShares MSCI ACWI ETF & 6.73 \\ \cline{2-4}
 & CL=F & Crude Oil Futures & 5.92 \\   \cline{2-4}
 & GLD & SPDR Gold Shares ETF & 2.80 \\ \cline{2-4} 
 & EEM & iShares MSCI Emerging Markets ETF & 2.18 \\  \hline

\multirow{1}{*}{Crypto} & BTC-USD & Bitcoin USD & 0.82 \\

\bottomrule
\end{tabular}
    }
\end{table}

The table predominantly features equity-related tickers, which corresponds with the sentiment analysis's goal to evaluate feelings toward significant equity sectors. Moreover, the analysis intriguingly links the global market sentiment not only with traditional equities but also with diverse asset classes including Bitcoin, bonds, and gold, showcasing a multifaceted approach to understanding market dynamics. Even if we only showcase in table \ref{tab:ticker_distribution} less than half of the tickers, the table underscores the complexity and interconnectedness of modern financial markets, with over five thousands tickers and multiple underlying types (equities, bonds, commodities, credit, currencies and even crypto-currencies).

% \begin{table}[!htbp]
% \centering
% \begin{tabular}{llc}
% \textbf{Region}                 & \textbf{Ticker} & \textbf{Percentage (\%)} \\ \hline
% \multirow{4}{*}{US}             & SPY             & 13.23                     \\
%                                 & \^GSPC           & 10.29                     \\
%                                 & ^TNX            & 4.25                      \\
%                                 & UUP             & 2.37                      \\ \hline
% \multirow{2}{*}{Asia}           & MCHI            & 2.78                      \\
%                                 & ^N225           & 1.97                      \\ \hline
% \multirow{5}{*}{Europe}         & EWG             & 1.37                      \\
%                                 & VGK             & 0.96                      \\
%                                 & EZU             & 0.58                      \\
%                                 & FEZ             & 0.07                      \\
%                                 & DAX             & 0.001                     \\ \hline
% \multirow{4}{*}{Emerging Countries} & ACWI       & 6.23                      \\
%                                 & CL=F            & 6.49                      \\
%                                 & EEM             & 2.07                      \\
%                                 & GLD             & 3.21                      \\
% \end{tabular}
% \caption{Distribution of tickers by region}
% \end{table}

\subsection{Ticker Market Reaction}
The algorithm classifies financial headlines based on their impact on stock prices using a quantile approach. Historical stock performance data is utilized to categorize the impact of news on the stock prices in a systematic manner.

For each stock ticker, denoted as \( T_k \), the algorithm operates in the following manner:

\begin{enumerate}
    \item \textbf{Historical Data Retrieval:} Acquire historical closing prices for the past five years, maintaining a rolling window approach—considering the history up to the publication date of the relevant financial headline.
    \item \textbf{Percentage Change Calculation:} Compute the daily percentage change in closing prices the day after the publication of the headline, denoted as \(\Delta P_{T_k}\).
    \item \textbf{Quantile Determination:} Determine the specific quantiles \( Q_{0.3, T_k} \) and \( Q_{0.6, T_k} \), representing the 30\% and 60\% thresholds, respectively, based on the historical price data.
    \item \textbf{Classification:} Assign a category to the impact of the news—positive (\(+1\)), negative (\(-1\)), or neutral (\(0\))—by comparing the calculated \(\Delta P_{T_k}\) against the quantiles.
\end{enumerate}

The classification function, \( C(T_k, \Delta P_{T_k}) \), is defined by the following:

\[
C(T_k, \Delta P_{T_k}) = 
\begin{cases} 
+1 & \text{if } \Delta P_{T_k} > Q_{0.6, T_k}, \\
-1 & \text{if } \Delta P_{T_k} < Q_{0.3, T_k}, \\
0 & \text{otherwise}.
\end{cases}
\]

\begin{figure}[htbp]
\begin{center} % This centers the entire picture horizontally in the page
\begin{tikzpicture}[>=Stealth]
    % Draw the main horizontal arrow with dotted line on the left and solid line on the right
    \draw[-, line width=1pt, dotted] (-5,0) -- (-4.5,0); % Dotted part
    \draw[->, line width=1pt] (-4.5,0) -- (5,0); % Solid part
    
    % Draw the vertical lines and labels
    \draw[-, line width=1pt] (-4.5,0.5) -- (-4.5,-0.5) node[below] {$t-1250$};
    \draw[-, line width=1pt] (2.5,0.5) -- (2.5,-0.5) node[below] {$t$};
    \draw[-, line width=1pt] (3.2,0.5) -- (3.2,-0.5) node[below] {$t+1$};
    
    % Draw the rectangles
    \fill[opacity=0.3, blue!50] (-4.5,0) rectangle (2.5,0.3);
    \fill[opacity=0.3, paleblue] (2.5,0) rectangle (3.2,0.3);
\end{tikzpicture}
\end{center}
\caption{Time ordering of the automatic labelling steps. The headline is published at time $t$, the historical price distribution is gathered from $t-5$ years to $t$ and the next day price return is from $t$ to $t+1$. }
\label{fig:time_step_process}
\end{figure}
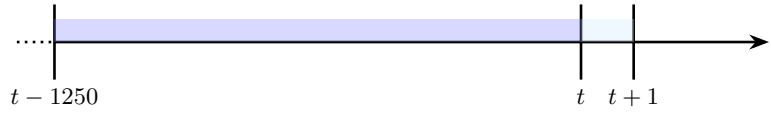

The sequence of steps over time plays a crucial role, as depicted in Figure \ref{fig:time_step_process}.

For a given headline published at time $t$, we examine the preceding five-year price history of the associated ticker, delineated in navy blue, up to the publication date. This historical percentage change is used to compare the subsequent day's return (from $t$ to $t+1$) against the historical distribution. This five-year span is chosen to align with our intent to evaluate the immediate effects of news while capturing a substantial portion of the ticker $T_k$'s past return distributions.

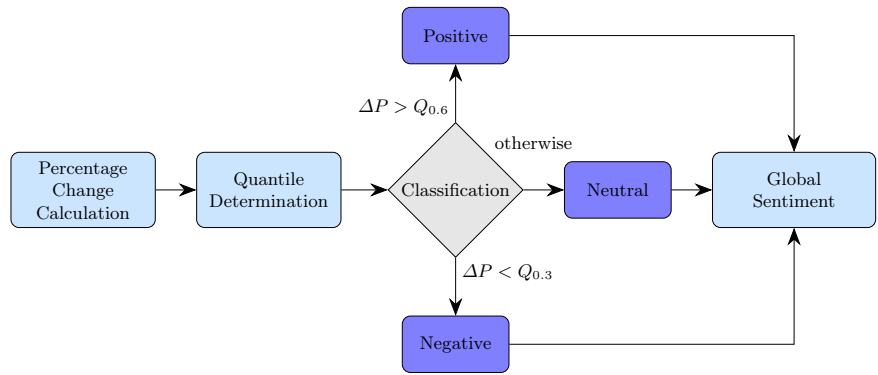
\begin{figure}[htbp]
\centering
\resizebox{0.95 \textwidth}{!}{% 
\begin{tikzpicture}[>=Stealth, 
    node distance=1.0cm and 0.7cm, % distance between nodes
    block/.style={rectangle, draw, fill=paleblue, text width=7em, text centered, rounded corners, minimum height=4em},
    label/.style={rectangle, draw, fill=blue!50, text width=5.0em, text centered, rounded corners, minimum height=3em},
    decision/.style={diamond, draw, fill=palegrey, text width=6.5em, text badly centered, node distance=0.8cm, inner sep=0pt},
    line/.style={draw, -{Stealth[length=3mm]}},
    auto]

    % Nodes
    \node[block] (calculation) {Percentage Change Calculation};
    \node[block, right=of calculation] (quantile) {Quantile Determination};
    \node[decision, right=of quantile] (classification) {Classification};
    \node[label, above=of classification] (positive) {Positive};
    \node[label, below=of classification] (negative) {Negative};
    \node[label, right=of classification] (neutral) {Neutral};
    \node[block, right=of neutral, text width=8em] (global) {Global \\ Sentiment};
    
    % Paths
    \path [line] (calculation) -- (quantile);
    \path [line] (quantile) -- (classification);
    \path [line] (classification) -- node [near start] {$\Delta P > Q_{0.6}$} (positive);
    \path [line] (classification) -- node [near start] {$\Delta P < Q_{0.3}$} (negative);
    \path [line] (classification) -- node [near start, above=0.6cm] {otherwise} (neutral);
    \path [line] (positive.east) -| (global.north);
    \path [line] (negative.east) -| (global.south);
    \path [line] (neutral) -- (global);
\end{tikzpicture}
}
\caption{Automatic Classification of Financial Headlines. Blue blocks represent data processing steps, green block represents decision points for classification, and orange blocks corresponding labels.}
\label{fig:label_process}
\end{figure}

This framework provides insight into news impact, utilizing historical volatility and performance benchmarks. The figure \ref{fig:label_process} summarizes the process and explain that we start from an initial calculation of the percentage change of the identified ticker used to determine a quantile and then convert this quantile into a sentiment score that is either 'Positive' if the percentage change exceeds the 60th percentile, 'Negative' if below the 30th percentile, and 'Neutral' otherwise.

\begin{figure}[!htbp]
\centering
\resizebox{0.95 \textwidth}{!}{ 
\begin{tikzpicture}[>=Stealth, node distance=0.7cm and 0.5cm, % Adjusted horizontal distance
    block/.style={rectangle, draw, fill=paleblue, text width=8em, text centered, rounded corners, minimum height=4em},
    newblock/.style={rectangle, draw, fill=blue!50, text width=8em, text centered, rounded corners, minimum height=4em},
    imageblock/.style={rectangle, draw, text width=8em, text centered, minimum height=3cm}, % Style for the image block
    line/.style={draw, -{Stealth[length=2mm]}},
    group/.style={rectangle, draw=blue, dashed, thick, inner sep=0.2cm, rounded corners}, % Reduced inner sep for narrower rectangle
    auto]
    
    % Nodes
    \node[block] (collect) {Collect News Headlines};
    \node[block, above right=0.2 cm and 0.5 cm of collect] (human_read) {Human Read};
    \node[block, right=of human_read] (human_interp) {Human \\Interpret};
    \node[block, right=of human_interp] (human_label) {Human \\Label};
    
    \node[newblock, below right=0.2 cm and 0.5 cm of collect] (ticker_id) {Identify Ticker};
    \node[newblock, right=of ticker_id] (ticker_compute) {Evaluate Ticker Return};
    \node[newblock, right=of ticker_compute] (ticker_label) {Machine \\Label};
    
    \node[block, below right=0.2 cm and 0.5 cm of human_label] (evaluate) {Fine-tune Model}; % Positioned at the extreme right, middle position
    
    % Grouping rectangle
    \node[group, fit=(human_read) (human_interp) (human_label)] {};
    
    % Image block for Machine Label
    \node[imageblock, below=of ticker_label] (image_block_machine) {\includegraphics[scale=0.20]{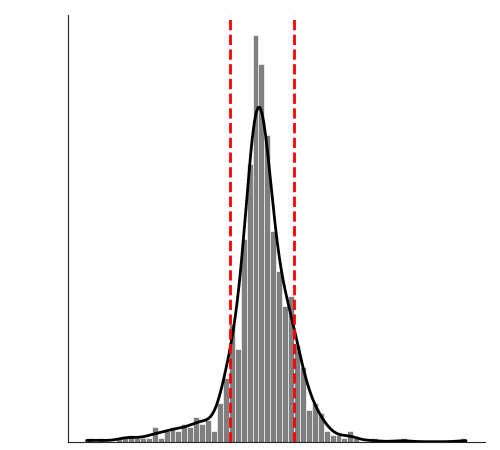}}; % Add your image here for Machine Label
    
    % Image block for Evaluate Ticker Return
    \node[imageblock, below=of ticker_compute] (image_block_evaluate) {\includegraphics[scale=0.2]{images/dist_resturn.png}}; % Add your image here for Evaluate Ticker Return
    
    % Paths
    \path [line] (collect) |- (human_read);
    \path [line] (human_read) -- (human_interp);
    \path [line] (human_interp) -- (human_label);
    
    \path [line] (collect) |- (ticker_id);
    \path [line] (ticker_id) -- (ticker_compute);
    \path [line] (ticker_compute) -- (ticker_label);
    
    \path [line] (human_label) -| (evaluate);
    \path [line] (ticker_label) -| (evaluate);
    
    % Connecting blocks to Image Blocks
    \path [line] (ticker_label) -- (image_block_machine);
    \path [line] (ticker_compute) -- (image_block_evaluate);
    
    % Additional nodes for the text labels on the arrows
    \node[above right=0.8 cm and -2.2cm of collect, align=center] (current_approach) {Current\\Approach};
    \node[below right=0.8 cm and -2.2 cm of collect, align=center] (new_approach) {New\\Approach};
\end{tikzpicture}
}
\caption{Full Process of Dataset Annotation.}
\label{fig:full_process_sum}
\end{figure}

Our approach of building an automatic annotator for sentiment classification is new to the financial sector. Unlike other industries, the initial step involves accurately labeling each news item, which is not a simple task. To summarize our contribution, we provide a recap of the entire process, including the new steps, as shown in figure \ref{fig:full_process_sum}.

The dotted light-blue box denotes the usual steps of annotating financial news in the sentiment classification task. They all presents human biases and are slow to achieve. The given labels might not be accurate and reflect the real global market reaction because of the human's sensitivity when reading. The red boxes are the new steps that proposes an alternative to avoid human biases and automate the annotation process. 

\subsection{Global Market-based Sentiment}
The global market-based sentiment of a headline is obtained by taking the median sentiment from its list of associated tickers. These sentiments are categorized as negative (\(-1\)), neutral (\(0\)), or positive (\(1\)). A headline can be linked to multiple tickers, and the prevailing sentiment across these tickers is considered the headline's global sentiment. The sentiment distribution in the dataset is balanced: 31\% negative, 27\% neutral, and 42\% positive.

\subsection{Dataset Validity Assessment}
After obtaining the dataset, we need to assess the validity of the labels. In order to ensure that the labels given to each headline are correct, we compare the stock market return with the return of a strategy based on the labels of the headlines. We developed a simple strategy that invests on the equity market according to the score's value. If the dataset is valid and thus the labels are consistent with the real-world market reaction to the headlines, the strategy's return must over-perform the stock market return. The closer the labels are to being random, the lower the returns will be in comparison to a random guess, whose performance is called in the following the benchmark. It is crucial to mention that if the headlines are accurately annotated, they should exhibit a look-ahead bias, considering that the annotations are made the day after the publication date.

For assessing the labeling of the dataset, we follow this process:
\begin{itemize}
    \item We generated a signal from the dataset based on the labels and following the sentiment score developed in \cite{lefort2024chatgpt}. 
    \begin{equation}
        S = \frac{\sum_{i=1}^{N} p(h_i) - \sum_{i=1}^{N} n(h_i)}{\sum_{i=1}^{N} p(h_i) + \sum_{i=1}^{N} n(h_i)} 
    \end{equation}
    \item We applied the signal on the equity market using the methodology introduced in \cite{lefort2024stress}. We used the signal on an equally weighted mix of the main equity markets. This also enabled us to compare it with a naive strategy that stands as the benchmark. If the strategy based on the sentiment signal is correct, then the induced look-ahead bias must outperform the benchmark. If the labeling is correct it means that the mapping between the headline and its next-day return must be correct.
\end{itemize}

\begin{figure}[!htbp]
    \centering
    {{\includegraphics[scale=0.5]{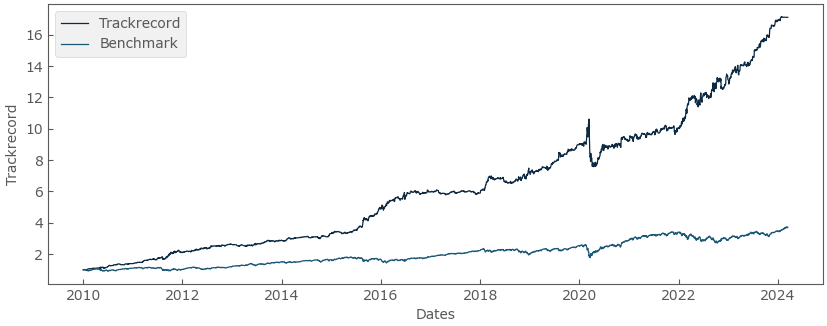} }}%
    \caption{Trackrecord of the labeling validity. The y-axis is the cumulative return of the strategies and the x-axis are the date of signal computation. The bigger the cumulative return is, the most efficient the classification is.}\label{fig:trackrecord_strat}%
\end{figure}

Figure \ref{fig:trackrecord_strat} shows clearly that the resulted strategy performs much better than the random guess benchmark. It therefore experimentally validate that the systematic annotation of our dataset is effective for predicting the equity market future reactions also referred to as the market sentiments in financial news analysis. It means that if we develop a model that can predict exactly the label in the dataset, the model would be able to detect the real-world impact of the headline on the market. In the case of a random sentiment signal, conversely,  we would obtain a flat, slightly downward curve. 

\section{Comparison of the Models on the Market-based Dataset} \label{sec:model_comparison}
We compare GPT-4 against two other open-source Large Language Models renowned for their accuracy in market reaction analysis within the financial domain: FinBERT and DistilRoBERTa, both specifically fine-tuned for financial news sentiment analysis. These models are accessible on the HuggingFace.co platform \cite{finbert_prosusai} and \cite{distilroberta_mrm8488}.

\subsection{Models Before Fine-tuning}\label{sec:model_before_finetuning}
Despite its considerably larger parameter count, GPT-4 exhibits only marginally better performance than the other models. The substantial difference in the number of parameters between GPT-4 and the more compact models does not translate into a significantly enhanced decision-making capability in analyzing financial texts, as shown in table \ref{tab:perf_before_finetuning}. This table highlights that, even though GPT-4 has advanced capabilities for financial text analysis, its improvement in decision-making is not markedly superior to that of BERT-based models. It is crucial to note that the F-score is weighted, prioritizing classes with a higher number of correct classifications.

\begin{table}[!htbp]
    \centering
    \caption{Model Performances On the 3-class Classification Before Fine-Tuning}
    \label{tab:perf_before_finetuning}
    \begin{tabular}{lcccc}
        \toprule
        \bfseries Model & \bfseries Param. Number & \bfseries Precision & \bfseries Recall & \bfseries F-score \\
        \midrule
        GPT-4 & $>1 \times 10^{11}$ & 0.48 & 0.49 & 0.47 \\
        DistilRoBERTa & $110 \times 10^{6}$ & 0.45 & 0.45 & 0.44 \\
        FinBERT & $82.1 \times 10^{6}$ & 0.46 & 0.46 & 0.44 \\
        Random & \ldots & 0.33 & 0.33 & 0.33 \\
        \bottomrule
    \end{tabular}
\end{table}

Challenges arise particularly with the neutral class, which all models struggle to classify accurately. This difficulty is attributed to the complex financial insights often embedded in headlines with a neutral market impact. Accurate classification in these cases demands deep financial market knowledge and experience, which the models lack. To assess the models' ability to develop this intricate reasoning, they were fine-tuned on a baseline dataset.

\subsection{Fine-tuned Models}\label{sec:fine_tuned_models}
To enable Large Language Models (LLMs) to uncover hidden patterns for prediction, we fine-tuned them using a market-based dataset, allocating 70\% of the dataset for training, which encompasses approximately 50,000 headlines. This represents a significant volume of data for model training. All models underwent fine-tuning with identical parameters and the same dataset. Furthermore, the evaluation set remained consistent across all models.

SFT refers to "Supervised Fine-Tuning", a process where models are trained on a specific task to improve their performances. Following this process, Large Language Models (LLMs) maintain similar levels of performance, with Table \ref{tab:perf_after_finetuning} illustrating that no model outperforms the others significantly. This observation indicates that the total parameter count of these models does not majorly influence their effectiveness in this specific classification task. The training time is also much higher with the increased number of parameters which appears as a disadvantage about the GPT model for the classification task.

\begin{table}[!htbp]
    \centering
    \caption{Fine-Tuned Model Performances}
    \label{tab:perf_after_finetuning}
    \begin{tabular}{lcccc}
        \toprule
        \bfseries Model & \bfseries Train. Time (hrs) & \bfseries Precision & \bfseries Recall & \bfseries F-score \\
        \midrule
        SFT GPT & 24 & 0.54 & 0.53 & 0.51 \\
        SFT DistilRoBERTa & 10 & 0.53 & 0.51 & 0.49 \\
        SFT FinBERT & 12 & 0.54 & 0.52 & 0.50 \\
        Random & \ldots & 0.33 & 0.33 & 0.33 \\
        \bottomrule
    \end{tabular}
\end{table}

\subsection{Performance Per Classes}
\begin{figure}[!htbp]
    \centering
    {{\includegraphics[scale=0.35]{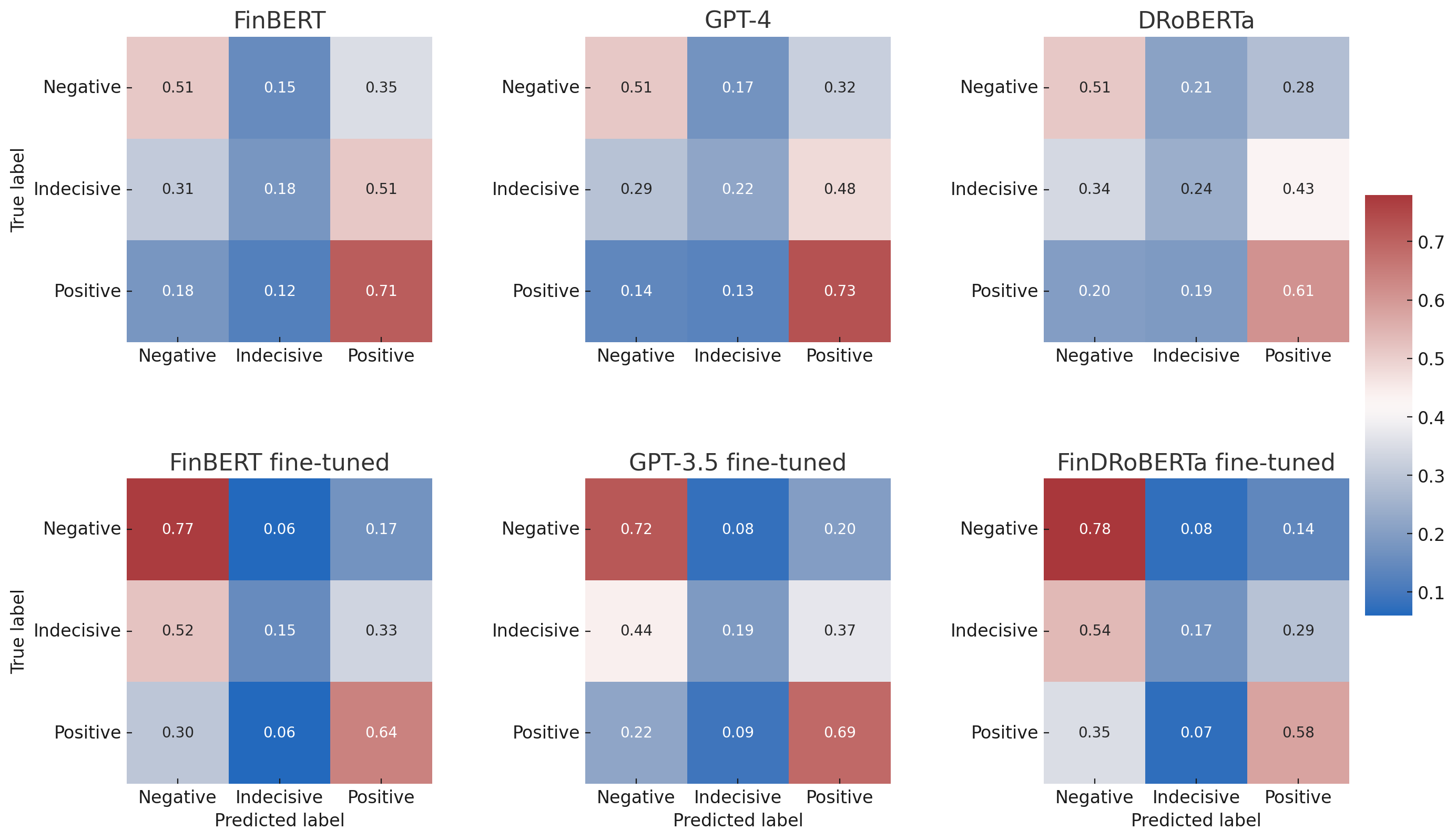} }}%
    \caption{Confusion matrices for each LLM on their ability to classify specific categories. }\label{fig:matrix_grid}%
\end{figure}

We explore whether specific large language models (LLMs) exhibit superior performance in predicting certain classes, as illustrated in figure \ref{fig:matrix_grid}. The premise is that if one model consistently outperforms others in a particular category, it would indicate a bias introduced during the training phase or when retrieving results. Additionally, this could also reflect a model's adeptness at interpreting positive or negative events, or a conservative tendency particularly effective in identifying the indecisive class. However, our findings reveal 
no single model consistently outperformed the others in every category, even after fine-tuning. Interestingly, post-fine-tuning, all models showed improved accuracy in predicting positive events, a reversal from their initial inclination towards negative events. This shift suggests that the original training on financial data lead to a negative bias, which was mitigated after retraining with our market-oriented dataset.

After observing how the model performs with certain categories, we considered that the overall performance might be enhanced by using a bagging method. The classifiers all supported the hypothesis of Condorcet's theorem: each performed significantly better than a random choice and comparable with each other. Also, they were assumed to be independent. However, we experienced that the hypothesis of independence could not hold, as the bagging model did not surpass the individual models, suggesting that all models make classification decisions in a similar manner, regardless of their size or number of parameters.

\subsection{Majority Vote Classifier and Non-Independence}\label{sec:MajorityVoteClassifier}
In this section, we propose a formal approach for demonstrating that the equal performance of the models on the classification task induces a common pattern of decision for all the LLMs. The effectiveness of the bagging method relies on the Condorcet's theorem that states that if several independent estimators, that are better than random decision, are combined into a majority vote classifier, the resulting classifier should have an improved performance and beat any of the individual models. Conversely, if the bagging method does not lead to improved performance, the independent hypothesis can not hold and individual models exhibit some dependencies in their behaviors.

The bagging method combines predictions from several base estimators by majority vote to improve the algorithm accuracy. Mathematically, the majority vote classifier selects the label with the most votes among $n$ base classifiers $\{C_1, C_2, \ldots, C_n\}$ mapping inputs from space $X$ to labels in space $Y = \{1, \ldots, k\}$. For all $n$ and $m \neq n$, $C_n$ is independent from $C_m$. The key assumption beyond independence is that all the  individual classifiers, $C_n$, are a better than random guessing. If multiple labels receive the same highest number of votes from classifiers, we take the average label, provided it does exist (for instance 0 if -1 and 1 are the highest classes). Otherwise, we take randomly one of the best labels. 

\subsection{Majority Vote Classification}\label{sec:Majority Vote}
Since all our individual models are better than random, the majority vote bagging method should perform better provided the independence between each individual models holds, as highlighted in section \ref{sec:MajorityVoteClassifier}. We selected the majority class given by an ensemble of LLMs that presents similar individual performances. To make this approach robust, we tried different bagging sets to avoid any selection bias with respect to the individual models in our experiment.
We provided in table \ref{tab:ensemble_performance} the performances of the resulting bagging model. Below is the list of the three different bagging configurations, including the SFT "Supervised Fine-Tuning" models.

\begin{samepage}
\begin{itemize}
    \item Bagging 1: SFT GPT + SFT DistilRoBERTa + SFT FinBERT.
    \item Bagging 2: SFT DistilRoBERTa + SFT FinBERT.
    \item Bagging 3: All the SFT and No SFT models.
\end{itemize}
\end{samepage}

\begin{table}[!htbp]
    \centering
    \caption{Ensemble Method Performances}
    \label{tab:ensemble_performance}
\resizebox{\textwidth}{!}{     
    \begin{tabular}{lccc}
        \toprule
        \bfseries Model & \bfseries Precision & \bfseries Recall & \bfseries F-score \\
        \midrule
        SFT GPT + SFT DistilRoBERTa + SFT FinBERT & 0.55 & 0.53 & 0.51 \\
        SFT DistilRoBERTa + SFT FinBERT & 0.53 & 0.52 & 0.52 \\
        All SFT and No SFT Models & 0.53 & 0.53 & 0.52 \\
        Random & 0.33 & 0.33 & 0.33 \\
        \bottomrule
    \end{tabular}
    }
\end{table}

The bagging experiment demonstrates that there is no significant improvement, regardless of whether the models are fine-tuned or not. Individual models significantly outperform random selection, showing comparably high performance levels. However, according to the premises of Condorcet's theorem, two conditions are required: the models must outperform random selection and they must be independent. Since the first condition is met, the failure to observe improvement suggests that the second condition is not fulfilled. This implies that the models cannot be considered mathematically independent, indicating that they exhibit similar behaviors. This finding is interesting as it highlights similarities between model of different parameters size.

\section{Conclusion and Future Works}\label{sec:Conclusion}
In this paper, we illustrate that with proper fine-tuning, compact, non-generative models such as FinBERT and FinDRoBERTa are able to  surpass the zero-shot learning capabilities of larger models like GPT-3.5 and GPT-4 in analyzing the sentiment of financial news. These results are interesting as they challenge the common assumption that larger generative models inherently possess superior performance in all contexts. To fine-tune models, we developed a new and large dataset derived from all major Bloomberg news from 2010 to 2024 that assigns market reaction to news headlines while eliminating human bias, thanks to systematic labelling. This dataset as well as the resulting models are available on HuggingFace. In addition, the application of a bagging method did not surpass the performance of the best individual models, thereby challenging the independence hypothesis of the Condorcet's theorem. This outcome suggests a similarity or at least some uniformity in decision-making patterns among large language models (LLMs). Future research should examine whether the efficiency of smaller, tailored models against larger counterparts like GPT\mbox{-}3.5 and GPT\mbox{-}4 extends to other datasets and tasks beyond the scope of financial sentiment analysis.

\bibliographystyle{splncs04}
\bibliography{main}
\end{document}